\documentclass{article}

\usepackage{arxiv}

\usepackage[utf8]{inputenc} 
\usepackage[T1]{fontenc}    
\usepackage{xurl}
\usepackage{hyperref}       
\usepackage[capitalise]{cleveref}
\usepackage{url}            
\usepackage{amsfonts}       
\usepackage{nicefrac}       
\usepackage{microtype}      
\usepackage{lipsum}		
\usepackage{graphicx}
\usepackage{natbib}
\usepackage{doi}
\usepackage[normalem]{ulem}
\usepackage[dvipsnames]{xcolor}
\usepackage{csquotes}

\definecolor{refcol}{RGB}{23, 63, 95}

\hypersetup{
    bookmarks=true,         
    unicode=false,          
    pdftoolbar=true,        
    pdfmenubar=true,        
    pdffitwindow=false,     
    pdfstartview={FitH},    
    pdftitle={YOLO - You only look 10647 times},    
    pdfauthor={Christian Limberg, Andrew Melnik, Augustin Harter, Helge Ritter},     
    pdfsubject={YOLO - You only look 10647 times},   
    pdfcreator={Christian Limberg},   
    pdfproducer={Christian Limberg}, 
    pdfkeywords={YOLO, You Only Look Once, Object Detection, Explainable AI}, 
    pdfnewwindow=true,      
    colorlinks=true,       
    linkcolor=refcol,          
    citecolor=refcol,        
    filecolor=refcol,         
    urlcolor=refcol        
}

\title{YOLO - You only look \sout{once} 10647 times}

\author{ \href{https://orcid.org/0000-0002-4903-3933}{\includegraphics[scale=0.06]{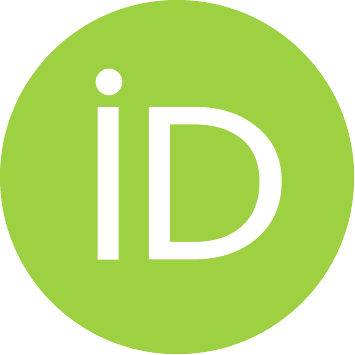}\hspace{1mm}Christian Limberg}\\
	Bielefeld University\\
	Germany \\
	\texttt{cnlimberg@gmail.com} \\
	\And
Andrew Melnik\\
	Bielefeld University\\
	Germany \\
	\texttt{andrew.melnik.papers@gmail.com} \\
	\AND
	Augustin Harter\\
	Bielefeld University\\
	Germany \\
	\texttt{aharter@techfak.uni-bielefeld.de} \\
	\And	
	Helge Ritter\\
	Bielefeld University\\
	Germany \\
	\texttt{helge@techfak.uni-bielefeld.de} \\
}

\renewcommand{\shorttitle}{YOLO - You only look 10647 times}

\begin{document}
\maketitle

\begin{abstract}
With this work we are explaining the \enquote{You Only Look Once} (YOLO) single-stage object detection approach as a parallel classification of 10647 fixed region proposals. We support this view by showing that each of YOLOs output pixel is attentive to a specific sub-region of previous layers, comparable to a local region proposal. This understanding reduces the conceptual gap between YOLO-like single-stage object detection models, RCNN-like two-stage region proposal based models, and ResNet-like image classification models. In addition, we created interactive exploration tools for a better visual understanding of the YOLO information processing streams: \href{https://limchr.github.io/yolo_visualization}{https://limchr.github.io/yolo\_visualization}
\end{abstract}


\section{Introduction}

Much progress in detecting multiple objects in an image using deep neural networks is related to the introduction of R-CNN \citep{girshick2014rich}, SSD \citep{liu2016ssd}, and YOLO \citep{redmon2016you} architectures. R-CNN and its improved versions like FasterRCNN \citep{fasterrcnn} detect objects by first producing proposals for regions containing an object \citep{melnik2021critic} and then in a second stage these proposals get passed through a classifier network. YOLO and SSD work without a separate proposal stage by combining object detection and classification into one stage. Four YOLO architecture successors along with other one-stage models were proposed, introducing improvements like the usage of anchor boxes, separate pathways for different object sizes, deeper architectures, different activation functions and a variety of other tricks and tweaks \citep{redmon2016yolo9000}, \citep{redmon2018yolov3}, \citep{bochkovskiy2020yolov4}.

We argue that the performance of these systems can be understood as performing classification for a high number of fixed region proposals with their positions relating to the convolutional grid.
To this end, we implement several interactive visualizations showing the inner processing of the network and providing detailed insights how YOLO achieves both high speed and high accuracy. As a core finding, using a saliency measure inspired from the Grad Cam approach \citep{selvaraju2017grad}, we show that each YOLO cell is attentive to a strongly localized sub-region of the image. Thus, we can understand YOLO as performing massively parallelized classification and regression of many image sub-regions positioned in a grid. These regions efficiently share most of their computation resulting from a high amount of overlap. We finally point out some analogies of this computational strategy to computational structures found in biological vision \citep{melnik2018world}, where a good trade off between speed and accuracy is tantamount as well.

\section{YOLO.v4 architecture}

We used a TensorFlow implementation\footnote{\url{https://github.com/hunglc007/tensorflow-yolov4-tflite}} of YOLO.v4~\citep{bochkovskiy2020yolov4}, that uses the original weights, in our experiments. Other YOLO implementations can be found here: YOLO.v1-v3\footnote{\url{https://pjreddie.com/darknet/yolo/}} and YOLO.v5\footnote{\url{https://github.com/ultralytics/yolov5}}. 

The network architecture is divided into two parts (see \cref{fig:yolo_arc}): First, the image is processed by a backbone network for feature extraction and second, the YOLO head is calculating object bounding boxes.

\subsection{Backbone network}

In YOLO.v4, darknet53 is used as a backbone for feature extraction. The network assumes input rgb-images of size $(416:416)$. It consists of 23 residual blocks and 77 convolutional layers. Five of the convolutional layers downsample the input by applying strides of 2. The backbone's output feature map and 2 intermediate outputs (after the 4. and 5. downsampling) are then passed into the YOLO.v4 head.

\subsection{YOLO head}

\begin{figure}
    \centering
    \includegraphics[width=\linewidth]{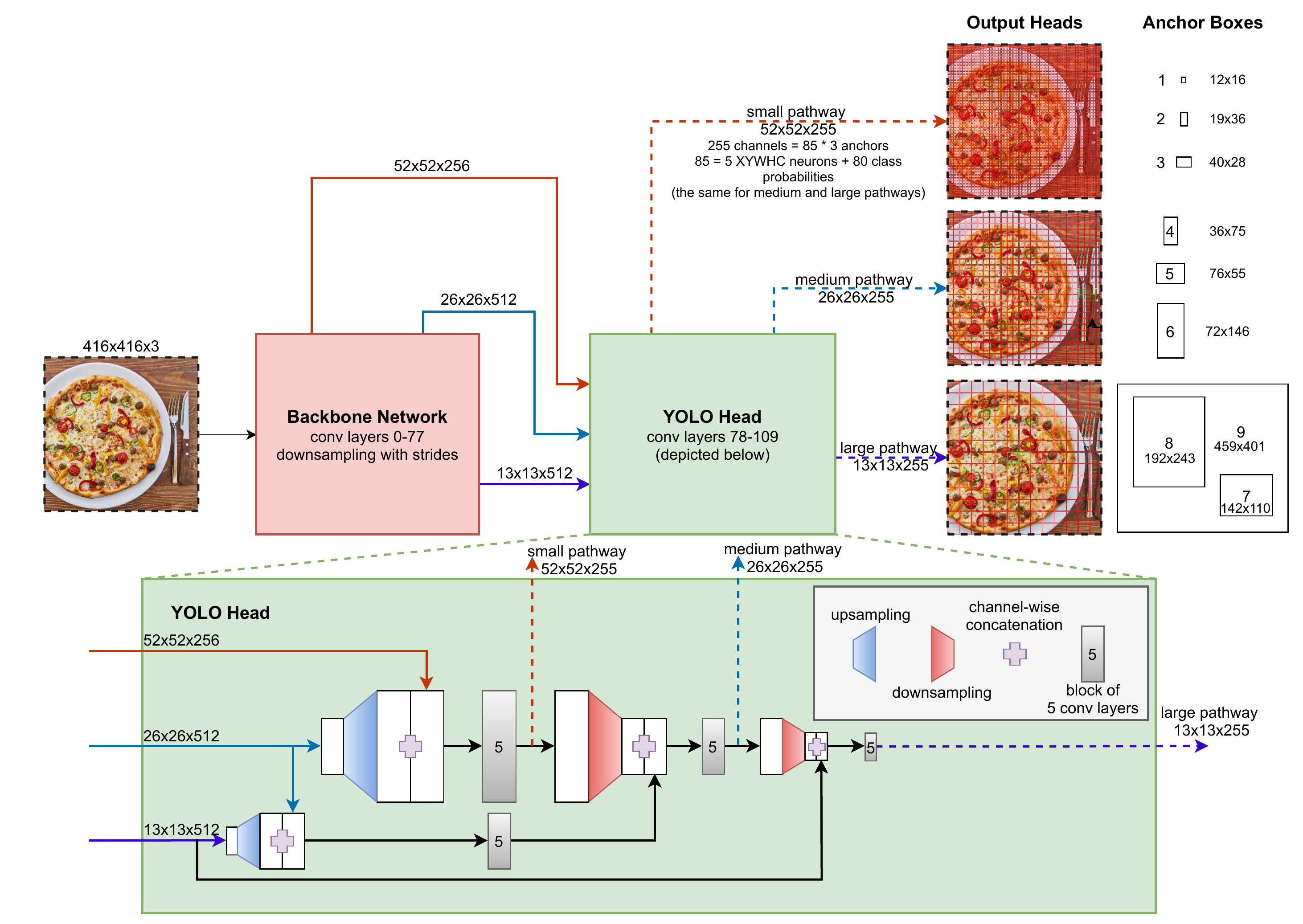}
    \caption{A simplified schematic of the YOLO.v4 network architecture.}
    \label{fig:yolo_arc}
\end{figure}

The YOLO head consists of 31 convolutions with strides 1 and padding for ensuring the same output size. Also, the three different paths from the backbone are concatenated in the YOLO head at different points with a combined upsampling and downsampling (see \cref{fig:yolo_arc}). 

The YOLO.v4 head has 3 different output \textit{pathways} supporting the detection of smaller, medium and larger sized objects.

The first output from the YOLO head is the small objects pathway after 2 upsampling operations and 2 concatenations with all paths of the backbone network. After another downsampling and concatenation with skip-connection, the medium objects pathway is defined and after a last downsampling paired with a concatenation, the final output of the large objects pathway is defined.

\subsection{Anchor boxes}

Anchor boxes are predefined bounding box patterns used by YOLO to delineate regions for object candidates.
Each of YOLO's three pathways uses 3 anchor box patterns (9 in total, see \cref{fig:yolo_arc} right column).
Each pathway provides for each of its output pixels 255 output channels (52x52x255, 26x26x255, and 13x13x255). 
Therefore, each anchor box has 85 channels (85=255/3). Five of these channels (XYWHC) represent the x- and y-displacement of the center of the object's bounding box, the width and height of the bounding box and a confidence value denoting that the specific anchor box is detecting an object. The remaining 80 channels are taylored to represent a probability value for each of the
80 classes for the COCO dataset.

For training, all bounding boxes from the COCO dataset were assigned to one (or several) of the YOLO anchor boxes, based on their IOU value with the box patterns (see section \ref{Training} for more details). Thus, after training, each of these 9 anchor boxes is trained to predict its own type of bounding box shape. Thus, the learnt anchor boxes work like templates for detecting objects of different sizes and shapes. There is e.g. an anchor box for detecting rather big, vertical-shaped objects and there is an anchor box for detecting smaller horizontal objects, etc. In total, the YOLO architecture provides us for every location in the input image with 9 bounding box proposals (3 pathways, each having 3 possible anchor box patterns). A confidence value near 0 indicates that no object was found located \enquote{under the output pixel}. In the training section (\ref{Training}) we further discuss how the training signal for the different anchor boxes, i.e. their representing neurons, is calculated.

\subsection{Training}\label{Training}

As explained in the previous section, the YOLO network is trained to map a 416x416x3 input array into three head output pathways (52x52x255, 26x26x255, and 13x13x255) to represent the input at appropriate discretization step (cell) for "large", "medium" and "small" objects, with each cell specifying three 85-dimensional channels for encoding up to three bounding boxes with contained objects. A cell can also opt not to represent its maximum of three boxes\footnote{note that the maximum of three representable boxes per channel is not in any deep way related to the number of three pathways - it is perfectly thinkable to specify YOLO architectures where these numbers are chosen to differ.}: each channel that represents a box indicates this by setting its C-variable to 1, thereby asserting the XYWH+classification meaning for the remaining 4+80 components. Otherwise, if C=0, these components are not to be interpreted as specifying anything and should all be set zero.

With this representation convention, any dataset with images of annotated axis-aligned bounding boxes that represent the outline of objects can be straightforwardly translated into target values for channel components of the three output pathways. For each image and each bounding box label, the 3 differently sized output pixels at the spatial center of the object are identified. Each output pixel can be represented by 3 different anchor boxes. For training the network, the anchor boxes that have an IOU greater than 0.3 compared to the annotated object bounding boxes from the dataset  are chosen to get a non-zero training signal. If there is no anchor box fulfilling this criteria, the anchor box with the largest IOU is selected to get a positive training signal. The anchor boxes with the positive training signal getting a confidence value of 1, encoded X,Y,W,H values of the respective object bounding box and a soft 1-hot-encoded vector that represents the object class. All other anchor boxes neurons are getting a training signal of 0.

With the so-defined target values the architecture can be straightforwardly trained to model the input-output relationship in the training data. This architecture can be compared to an ensemble of output paths, where each path is trained to detect the most appropriate bounding boxes \citep{bach2020error}.

\section{Visualization}

\begin{figure}
    \centering
    \includegraphics[width=0.8\linewidth]{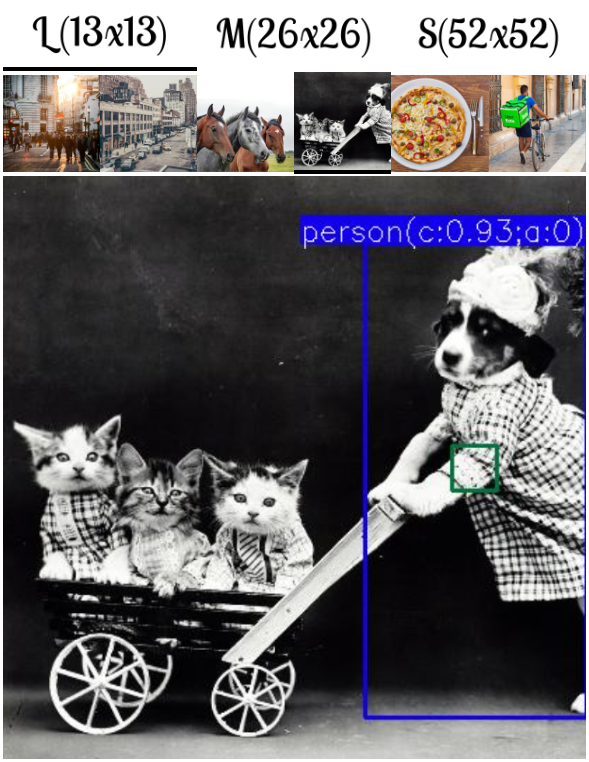}
    \caption{With our interactive visualization, the full grid layers of the YOLO.v4 network can be depicted for several images. The YOLO architecture has 3 different pathways for recognizing objects of different sizes. The recognition heads are located in 2d-grids of different resolutions. Each grid element can detect underlying objects based of 3 possible anchor box shapes. Each anchor box refines estimates of the x- and y-position, the width and the height, a confidence value and a probability vector of each class used for training. The bounding boxes are labeled with the predicted class, the certainty value and an index of the displayed anchor box (we depict only the most confident anchor box out of the 3 possible). Object proposals with a high certainty are colored blue.
    The interactive version of this plot can be accessed via \url{https://limchr.github.io/yolo_visualization/index.html\#fig2}.
}
    \label{fig:fig2}
\end{figure}

For a deeper inquiry into the nature of the pathway/output pixel representation that emerges
under this training in YOLO, we implemented an interactive visualization of all pathways/output pixels for several images (see \cref{fig:fig2}).

\begin{figure}
    \centering
    \includegraphics[width=0.8\linewidth]{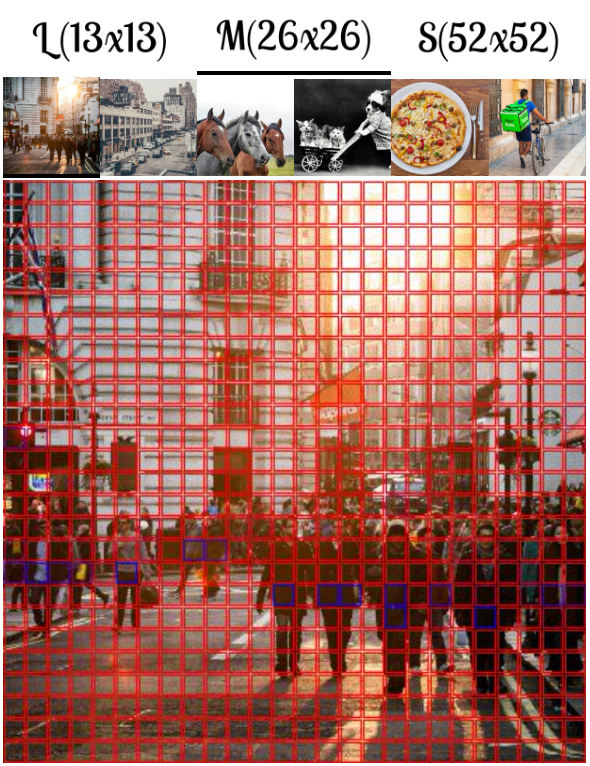}
    \caption{Shifting the actual input image below the output pixels makes apparent how the confidences of the anchor boxes (blue means high confidence) are shifted and neighboring cells get activated. 
    The interactive version of this plot can be accessed via \url{https://limchr.github.io/yolo_visualization/index.html\#fig3}.
}
    \label{fig:fig3}
\end{figure}

Since the set of non-zero channel cells is rather sparse, this also reduces the number of active (or high confident) output pixels for an object. \cref{fig:fig3} depicts this. By shifting the input image a few pixels, one can see how the output pixel's confidences are also shifting and a neighboring output pixel \enquote{takes over}. 

We find that a maximum of 4 output pixels have a high confidence to detect a certain object, but most of the time only 1 or 2 output pixels are active. \cref{fig:fig2} further shows, that all of these active output pixels have fairly similar bounding boxes, so as a post-processing step an ordinary non-maximum suppression can be applied, choosing the bounding box with the highest confidence.

\subsection{Saliency-based analysis of YOLO Layers}

In this section we present an analysis of the responses of YOLO's output pathways in terms 
of 
their \enquote{saliency pattern} of an intermediate convolutional layer. To compute a suitable saliency measure,
we build on the Grad Cam approach~\citep{selvaraju2017grad} which calculates gradients based on an output neuron that is under consideration (e.g., typically a class-neuron of the one-hot encoded output layer of a classification CNN). 

To produce a scalar that indicates the importance of a particular feature channel, the original algorithm first averages the gradients at the (usually last) convolutional layer over the spatial dimensions of the layer's channels. The result is then multiplied with the actual activation values of the neuron to get an importance weighted saliency map. The spatial pattern of these values across the convolutional layer and, therefore the corresponding input image, show which input areas are responsible for the response at the selected output neuron.

For applying this technique to YOLO, we have to make several changes. We can not use the very last convolutional layer because of the different architecture of the network (the last convolutional layers don't have that much saliency information because of the grid outputs). Therefore, we use an intermediate layer for computing saliency maps.

\begin{figure}
	\centering
	\includegraphics[width=1\linewidth]{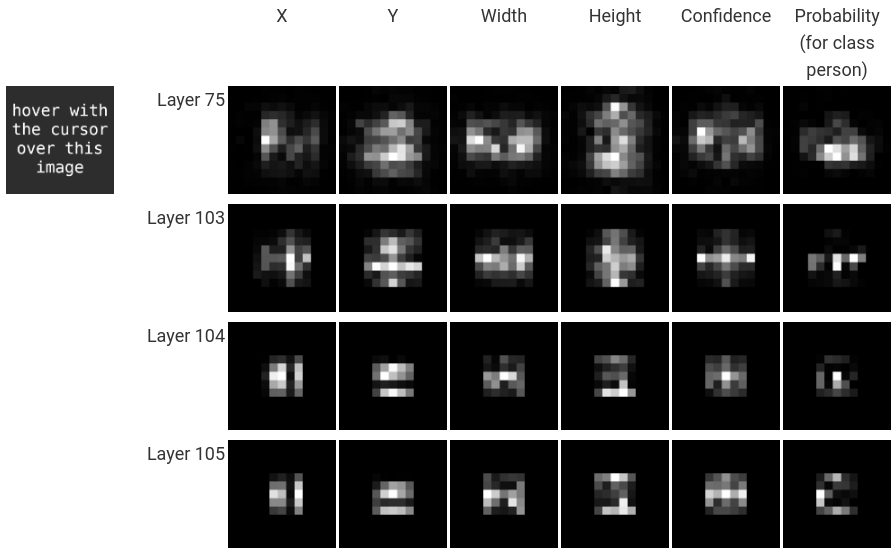}
	\caption{Our adapted Detection Grad Cam visualizes the saliency map of a single output neuron. The columns represent the saliency maps of the x-shift, y-shift, width, height, confidence and probability neuron of a selected output pixel of the large pathway (13x13 output pixels). As rows, we depict saliency maps for convolutional layers 75, 103, 104 and 105. Each plot represents the saliency map averaged over 15 images (15x13x13) having class \enquote{person} under the corresponding output pixel's position. 
		The interactive version of this plot can be accessed via \url{https://limchr.github.io/yolo_visualization/index.html\#fig4}.
	}
	\label{fig:fig4}
\end{figure}

We found out, that we can generate much more descriptive saliency maps by multiplying the output of the intermediate layer element-wise with the plain gradients of this layer (e.g. calculated from the w-neuron or the h-neuron of one specific output pixel/anchor box) and average the output channels. But still, the so-obtained result is pretty noisy. We did another trick for getting a clearer pattern: We query several images from the COCO dataset that have an instance of a particular class (e.g. \enquote{person}) located at a particular spatial image position, i.e. where the underlying YOLO output pixel would get a positive training signal. We pass 15 of these images through our adapted Grad Cam algorithm and average the result images for getting an saliency map for this output pixel. By repeating the process for all output pixels (omitting border output pixels since in the dataset are too few samples having persons located in the border areas), we get a clearer visualization and we can see how the saliency of YOLO will shift as we hover over the image (see \cref{fig:fig4}).

The Figure shows, that the saliency is focused below the corresponding grid-cell’s position. Further, it depicts that the saliency map of the w-neuron and x-neuron has a rather wider activation, while the h-neuron's and y-neuron's saliency map has a rather vertical activation i.e. the detection is more sensitive to these areas. The activation of the p-neuron and especially the c-neuron is more focused to the center.

\section{Discussion}

The visualization of the saliency patterns of YOLO neurons makes apparent that YOLO acts like parallelized classification CNNs: each anchor box's saliency is pointed to an underlying subarea of the image and on this subarea classification and regression tasks are focused. These tasks are locally interdependent (the dimensions and location of a box are determined by its contents and vice versa). However, this tight spatial coupling focuses the propagation of the gradients and their interactions, creating a strong \enquote{spatial hierarchy}-bias that makes learning and subsequent processing very efficient.

This is not dissimilar to the information processing in human visual cortex. In the primary visual cortex V1, features are extracted and passed to secondary visual cortex V2, where the information is split into a dorsal and ventral stream for localizing (dorsal) and classifying (ventral) objects. The dorsal stream is more explorative, showing a wider activation in the biological saliency map for localizing objects and movements in the scenery, where the ventral stream's activation focuses more on the center of the object (see \cite{sheth2016two} Fig. 2). 

We can see similar properties also from \cref{fig:fig4} when comparing columns x,y,w,h with c,p: x,y define the relative position of the detected object and w,h represent the dimensions of the object. I.e. the four neurons estimate where the object is, while c and p determine if there is an object present, and the object's class (what is it?). The saliency map activations of x,y,w,h is rather wide, focusing on the object borders (x,w for horizontal and y,h for vertical border areas) for determining where exactly it is, and the activation of c,p is more narrow and focused to the center, comparable to the dorsal and ventral stream of the biological model.

For YOLO.v4, the detection of a picture leads to a total number of 10647 classification/regression pairs, obtained by summing up the squared grid shapes times the number of anchor boxes ($(13^2 + 26^2 + 52^2) * 3$). This number is dependent on the image dimensions used for train YOLO.v4 which is 416x416 pixels. It has to be adapted for other image dimensions, since the dimensions of the grids would also change, if we assume the same network architecture.

The nature of an artificial neural network, and especially for a CNN, is that it consists of many parallel operations for one layer. This relates to a simple matrix multiplication. GPUs are built for this purpose and modern GPUs can do many matrix calculations in a very short amount of time. So it seems natural to exploit this feature and just \enquote{throw away the uninteresting results} i.e. the low confidence detections.

Compared to early 2-stage detectors, which had intermediate steps for selecting e.g. region proposals which were then fed into a second neural network, the advantage is a massive speed increase (and only a minor loss in accuracy) since this intermediate steps take time since they are running on the CPU. 

Our key message is that YOLO is not really "looking once", but 10647 times. Because of a clever exploitation of Artificial Neural Network structures, which make it possible to share most of the computation between regions and also allow to easily parallelize the computations on a GPU, this can be very fast and efficient.

\newpage
\bibliographystyle{unsrtnat}
\bibliography{paper} 

\end{document}